\documentclass[journal]{IEEEtran}

\usepackage{graphicx}
\usepackage{amsfonts}
\usepackage{amsmath}
\usepackage{booktabs}
\usepackage{graphicx}
\usepackage{cite}
\usepackage{picinpar}
\usepackage{url}
\usepackage{flushend}
\usepackage{colortbl}
\usepackage{soul}
\usepackage{multirow}
\usepackage{pifont}
\usepackage{color}
\usepackage{alltt}
\usepackage{enumerate}
\usepackage{siunitx}
\usepackage{breakurl}
\usepackage{epstopdf}
\usepackage{pbox}
\usepackage[hidelinks]{hyperref}
\usepackage{caption}
\usepackage{subcaption}
\usepackage[export]{adjustbox}

\usepackage{cite}

\ifCLASSINFOpdf
\else
\fi
\begin{document}
%
    
\title{A Transferable Adaptive Domain Adversarial Neural Network for Virtual Reality Augmented EMG-Based Gesture Recognition}
%
%
%

\author{Ulysse Côté-Allard, Gabriel Gagnon-Turcotte, Angkoon Phinyomark, \\ Kyrre Glette, Erik Scheme$\dagger$, François Laviolette$\dagger$, and~Benoit Gosselin$\dagger$
\thanks{$\dagger$ These authors share senior authorship}}
\maketitle

\begin{abstract}

Within the field of electromyography-based (EMG) gesture recognition, disparities exist between the offline accuracy reported in the literature and the real-time usability of a classifier. This gap mainly stems from two factors: 1) The absence of a controller, making the data collected dissimilar to actual control. 2) The difficulty of including the four main dynamic factors (gesture intensity, limb position, electrode shift, and transient changes in the signal), as including their permutations drastically increases the amount of data to be recorded. Contrarily, online datasets are limited to the exact EMG-based controller used to record them, necessitating the recording of a new dataset for each control method or variant to be tested. Consequently, this paper proposes a new type of dataset to serve as an intermediate between offline and online datasets, by recording the data using a real-time experimental protocol. The protocol, performed in virtual reality, includes the four main dynamic factors and uses an EMG-independent controller to guide movements. This EMG-independent feedback ensures that the user is in-the-loop during recording, while enabling the resulting \textit{dynamic dataset} to be used as an EMG-based benchmark. The dataset is comprised of 20 able-bodied participants completing three to four sessions over a period of 14 to 21 days. The ability of the dynamic dataset to serve as a benchmark is leveraged to evaluate the impact of different recalibration techniques for long-term (across-day) gesture recognition, including a novel algorithm, named TADANN. TADANN consistently and significantly (p$<$0.05) outperforms using fine-tuning as the recalibration technique.


\end{abstract}


%
\IEEEpeerreviewmaketitle

\section{Introduction}
Muscle activity as a control interface has been extensively applied to a wide range of domains from assistive robotics~\cite{applications_EMG} to serious gaming for rehabilitation~\cite{serious_game_rehabilitation} and artistic performances~\cite{dance_journal_paper}. This activity can be recorded non-invasively through surface electromyography (sEMG), a widely adopted technique both in research and clinical settings~\cite{applications_EMG, emg_survey_2007}. Intuitive interfaces can then be created by applying machine learning on the sEMG signal to perform gesture recognition~\cite{toon_emg_machine_learning}.

Despite decades of research in the field~\cite{emg_used_for_decades}, an important gap still exists between offline classifiers' performances and real-time applications' usability~\cite{toon_emg_machine_learning}. This disconnect stems in large part from the difficulty of including, within an offline dataset, the four main dynamic factors~\cite{transcientChangeEMG_including_gesture_intensity}: gesture intensity, limb position, electrode shift, and the transient nature of the EMG signal. This difficulty originates from the important time investments required to record even a fraction of the permutations which arise from these dynamic factors. Another point of concern is that creating more extensive recording protocols could be increasingly fatiguing for the participants~\cite{limb_orientation_plus_force_variation_offline}. Consequently, and to the best of the authors' knowledge, an offline dataset including all four dynamic factors simultaneously does not currently exist. 
At this point, it is important to highlight the fact that despite the associated difficulties, several studies have investigated the effect of these dynamic factors within an offline setting either individually (e.g. limb position~\cite{limb_position_study_2017}, electrode shift~\cite{electrode_shift_study_2016}, gesture intensity~\cite{force_variation_offline_with_vf_emg_channels}, transient change~\cite{user_adaptation_emg_even_without_feedback}) or as a subset~\cite{multiple_dynamic_factor_offline, limb_orientation_plus_force_variation_offline}. Further, it was shown that including these dynamic factors within the training dataset improves the robustness of the classifier~\cite{transcientChangeEMG_including_gesture_intensity, improve_performance_limb_position_in_training_offline}. These types of studies are of the utmost importance to understand how these factors affect the EMG signal and therefore how to create more robust controllers. However, in practice, it would not be desirable if the user had to record an extensive dataset before being able to use the system~\cite{transcientChangeEMG_including_gesture_intensity}. As such, for real-world use, the system should ideally be able to learn to contend with untrained conditions. 
Additionally, generally when recording offline datasets, participants cannot regulate their gestures through external feedback due to the absence of a controller. Therefore, users are forced to rely on internal focus (i.e. focus on their own movements/muscle activity), which affects their behavior, and consequently their EMG signals~\cite{emg_internal_vs_external_focus, feedback_vs_no_feedback_affect_EMG}. This effect might be particularly noticeable when the participant needs to record multiple repetitions over a long period (user learning) as would be the case if one wants to record an offline dataset including all four main dynamic factors. Contrarily, online myoelectric control naturally provides external feedback to the participant. In turn, this feedback biases the recorded online dataset towards the algorithm used for control, as the participants will adapt their behavior in an effort to improve the system's usability~\cite{emg_subject_learning_non_intuitive_control, longterm_amputee_adapt_overtime_with_feedbacks, journal_paper_TL_ulysse}. Consequently, the recording of a new dataset for each control method or variant to be tested is required for a fair comparison. Recording such datasets, however, is not only time-consuming but can also require expensive hardware (e.g. prosthetic arm, robotic arms)~\cite{robotic_arm_iros}. A common alternative to using this costly equipment are computer simulations (e.g. Fitts' law test~\cite{fitt_law_test_example}) running on a 2D computer screen. Unfortunately, these types of simulations limit the number of degrees of freedom that can be intuitively controlled. In contrast, virtual reality (VR) offers an attractive and affordable environment for EMG-based real-time 3D control simulations~\cite{emg_vr_2016}.

As such, one of this work's contributions is the creation of a virtual reality environment from which a \textit{dynamic dataset}, featuring 20 participants and recorded specifically to contain the four main dynamic factors, is made publicly available. An important innovation of this dataset is that the real-time, gesture recognition feedback is provided solely by a Leap Motion camera~\cite{leap_motion_product} (stereo camera). This dynamic dataset thus serves as an intermediary between an offline and an online dataset. That is, the experimental protocol was created to contain the four main dynamic factors and to also provide external feedback to the participant without biasing the dataset towards a particular EMG-based controller. The dataset can therefore be used as a benchmark to compare new EMG-based algorithms. In addition, the VR environment, in conjunction with the Leap Motion, smoothly tracks the participant's limb orientation in 3D, which provides a more precise understanding of the effect of limb position (compared to the pre-determined positions generally used when investigating this factor in the field~\cite{five_discrete_limb_position_emg_amputees, 16_position_limb_position_emg, limb_position_emg_at_different_specific_angles}). Further, the Leap Motion allows the VR environment to easily and intuitively provide feedback (both continuous and discrete) to the participant independently of (or in conjunction with) any potential EMG-based controller. For each user, the recorded data contains between three to four recording sessions (equally distant) spanning a period of 14 to 21 days. The recording sessions were ``gamified'' within the VR environment, to better engage the participants.

Using the dynamic dataset, this work proposes an analysis of the effect of the four main dynamic factors on a deep learning classifier. The feature learning paradigm offered by deep learning allows the classifier to directly receive the raw sEMG data as input and achieve classification results comparable with the state of the art~\cite{journal_paper_TL_ulysse, raw_emg_good}, something considered ``impractical'' before~\cite{emg_survey_2007}. This type of input (raw EMG) can be viewed as a sequence of one-dimensional images. While ConvNets have been developed to encode spatial information, recurrent neural network-based (RNN) architectures have been particularly successful in classifying sequences of information~\cite{deep_learning}. Hybrid architectures combining these two types of networks are particularly well suited when working with sequences of spatial information\cite{1_convNetRNN, 2_convNetRNN}. In particular, such hybrid networks have been successfully applied to sEMG-based gesture recognition~\cite{emgConvNetLSTM}. Temporal Convolutional Networks (TCN)~\cite{wavenet, originalTCN} on the other hand are a purely convolutional approach to the problem of sequence classification. Compared to the hybrid ConvNet-RNN, TCNs are parallelizable, less complex to train, and have low memory requirements. Within the context of real-time sEMG-based gesture recognition, especially if applied to prosthetic control, these computational advantages are particularly important. Additionally, TCNs have been shown to outperform RNN-based architectures in a variety of domains and benchmarks using sequential data~\cite{PyTorchTCN_implementation}. Consequently, this work proposes leveraging a TCN-based architecture to perform gesture recognition.

Another contribution of this work is a new transfer learning algorithm for long-term recalibration, named Transferable Adaptive Domain Adversarial Neural Network (TADANN), combining the transfer learning algorithm presented in~\cite{transfer_learning_conference_emg, journal_paper_TL_ulysse} and the multi-domain learning algorithm presented in~\cite{HandCraftVsLearnedFeaturesEMG}. The ability of the dynamic dataset to be used as a benchmark is leveraged to compare TADANN against fine-tuning~\cite{origin_fine_tuning}, arguably the most prevalent transfer learning technique in deep learning~\cite{fine_tuning_2012, fine_tuning_2014, fine_tuning_2016}

This paper is divided as follows; the VR experimental protocol and environment is first presented in Section~\ref{experimental_protocol}. Section~\ref{deep_learning_classifiers} then presents the deep learning classifiers and transfer learning method used in this work. Finally, the results and the associated discussion are covered in Section~\ref{results} and~\ref{discussion} respectively. 


\section{Long-term sEMG Dataset}
\label{experimental_protocol}
\begin{figure}[!ht]
\begin{center}
\includegraphics[width=\linewidth]{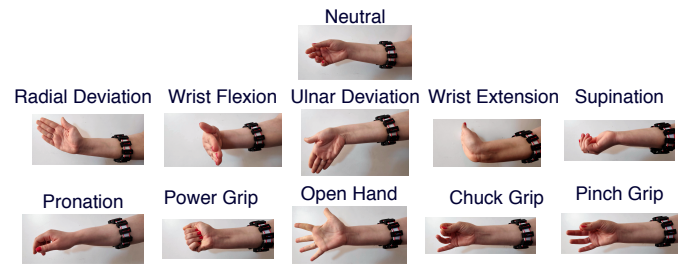}
\end{center}
\caption{The eleven hand/wrist gestures recorded in the \textit{Dynamic Dataset} (image from~\cite{3dc_armband})} 
\label{fig_gestures_dataset}
\end{figure}

This work provides a new, \href{https://github.com/UlysseCoteAllard/LongTermEMG}{publicly available (https://github.com/UlysseCoteAllard/LongTermEMG)}, multimodal dataset to study the four main dynamic factors in sEMG-based hand gesture recognition. The Dynamic Dataset features 20 able-bodied participants (5F/15M) aged between 18 and 34 years old (average 26 $\pm$ 4 years old) performing the eleven hand/wrist gestures depicted in Figure~\ref{fig_gestures_dataset}. For each participant, the experiment was recorded in virtual reality over three sessions spanning 14 days (see Section~\ref{vr_session_recording} for details). In addition to this minimum requirement, six of them completed a fourth session, so that the experiment spanned 21 days. 17 of the 20 participants had no prior experience with EMG systems, while the other three only had limited experience. Additionally, for 18 of the 20 participants, this was their first time within a VR environment. 
Note that originally, 22 persons took part in this study. However, two of them (both male) had to drop out, due to external circumstances. Consequently, these individuals are not included in the results and analysis of this work.

The data acquisition protocol was approved by the Comités d'Éthique de la Recherche avec des êtres humains de l'Université Laval (approbation number: 2017-026 A2-R2/26-06-2019). Informed consent was obtained from all participants.

\subsection{sEMG Recording Hardware}
\label{3dc_armband_section}
\begin{figure}[!htbp]
\centering
\includegraphics[width=.8\linewidth]{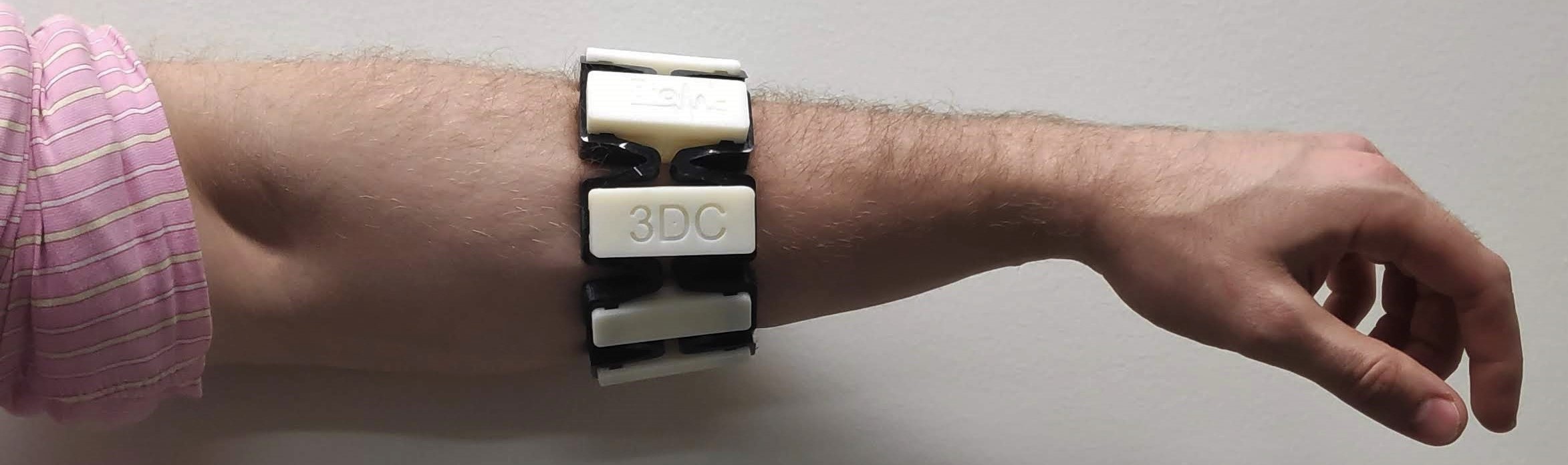}
\caption{The 3DC Armband used in this work records electromyographic and orientation (9-axis Magnetic, Angular Rate, and gravity sensor) data. The wireless, dry-electrode armband features 10 channels, each cadenced at 1~kHz.}
\label{fig_3DC_armband}
\end{figure}

The electromyographic activity of each participant's forearm was recorded with the 3DC Armband~\cite{3dc_armband}; a wireless, 10-channel, dry-electrode, 3D printed sEMG armband. The device, which is shown in Figure~\ref{fig_3DC_armband}, samples data at 1000 Hz per channel, thus covering the full spectra of sEMG signals~\cite{emg_200_vs_1000Hz}. In addition to the sEMG acquisition interface, the armband also features a 9-axis Magnetic, Angular Rate, and Gravity (MARG) sensor cadenced at 50 Hz. The dataset features the data of both the sEMG and MARG sensors at 1000 and 50 Hz respectively for each session of every participant.

\subsection{Stereo-Camera Recording Hardware}
During the experiment, in addition to the 3DC Armband, the Leap Motion camera~\cite{leap_motion_product} mounted on a VR headset was also used for data recording. The Leap Motion (https://www.ultraleap.com/) uses infrared emitters and two infrared cameras~\cite{leap_motion_details} to track a subject's forearm, wrist, hand, and fingers in 3D. In addition to the software-generated representation of the hand, the \textit{Dynamic Dataset} also contains the raw output of the stereo-camera recorded at $\sim$10 Hz.

\subsection{Experimental Protocol in Virtual Reality}
\label{vr_session_recording}

Recording the Dynamic Dataset within the developed VR environment, in conjunction with the Leap Motion camera, provided the following advantages in addition to the lower barrier of entry (compared to a robotic arm) and reproducibility: (1) Enables the software to intuitively communicate gesture intensity and position (in 3D) to the participant via the graphic interface. (2) The arm of the participant was replaced within the VR environment by a \textit{virtual prosthetic} providing direct, intuitive feedback (gesture detected, intensity, and position) to the participant. (3) Allows the experimental protocol to be easily gamified, which greatly aids both recruitment and participant retention. Finally, (4) the Leap Motion, in conjunction with a picture-based convolutional network (see Section~\ref{leap_motion_convNet_section}), is used as the real-time controller to provide feedback to the user without biasing the dataset toward a particular EMG-based classifier. Note that synchronization between the different modalities is managed using the timestamps recorded alongside each data-point for each modality and is provided as part of the published dataset.

\begin{figure}[!htbp]
\centering
\includegraphics[width=.9\linewidth]{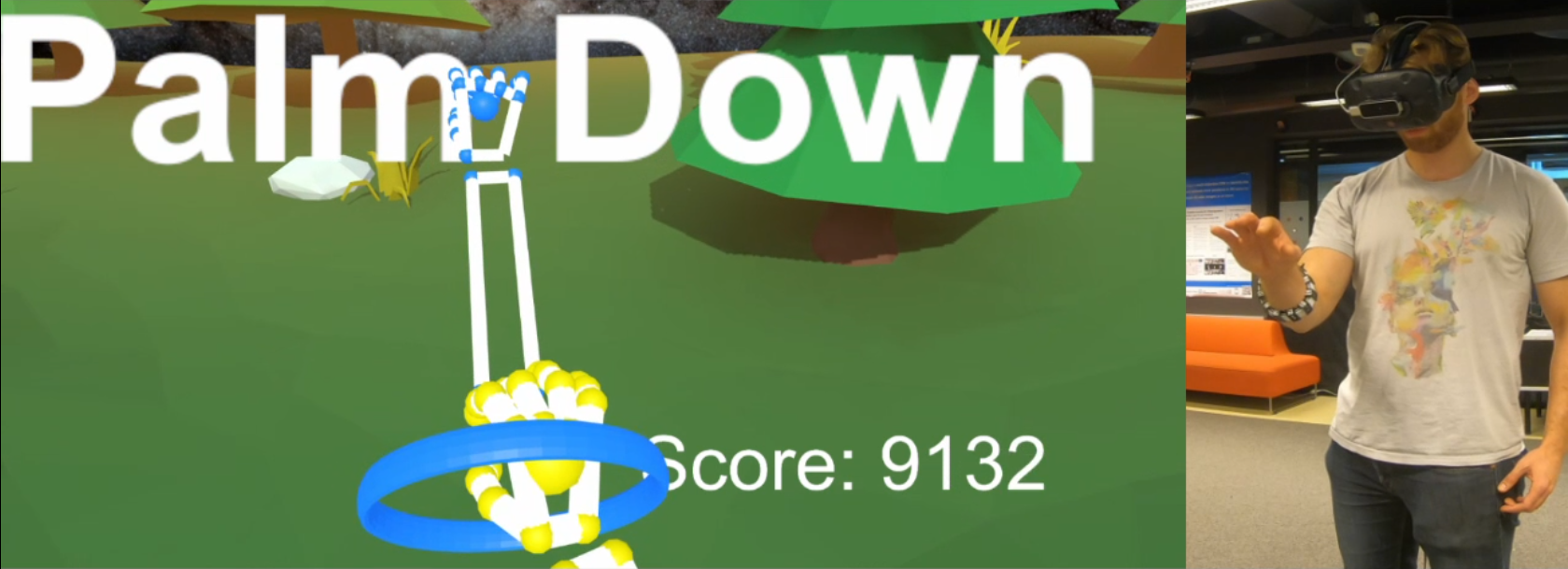}
\caption{The VR environment during the evaluation session. The scenery (trees, horizon) helps orient the participants. The requested gesture is written on the participant's head-up display and shown as an animation (the blue hand model). The ring indicates the desired hand's position while its color (and the color of the blue hand) indicates the requested gesture's intensity. The yellow hand represents the participant's virtual prosthetic hand and changes color based on the intensity at which the participant is performing the gesture. The score increases if the participant is performing the correct gesture. Bonus points are given if the participant is performing the gesture in the right position and intensity. Note that the software's screenshot only shows the right eye's view and thus does not reflect the depth information seen by the participant.}
\label{vr_interface}
\end{figure}

Each recording session is divided into two parts: the \textit{Training Session} and the \textit{Evaluation Session}, both of which are conducted in VR. Figure~\ref{vr_interface} helps visualizes the general interface of the software while \href{https://youtu.be/BnDwcw8ol6U}{the accompanying video (https://youtu.be/BnDwcw8ol6U)} shows the experiment in action. Note that for every training session, two evaluation sessions were performed. All three sessions were recorded within the timespan of an hour.

Before any recording started, the 3DC Armband was placed on the dominant arm of the participant. The armband was slid up until its circumference matched that of the participant's forearm. A picture was then taken to serve as a reference for the armband placement. In subsequent sessions, the participant placed the armband on their forearm themselves, aided only with the reference picture. Hence, electrode displacement between sessions is expected.

\subsubsection{Training Session}
\label{training_session_section}
The training session's main purpose was to generate labeled data, while familiarizing the participants with the VR setup. To do so, the participants were asked to put on and adjust the VR headset to maximize comfort and minimize blurriness. The VR platform employed in this work is the Vive headset (https://www.vive.com). After a period of adjustment of a few minutes, the recording started. All in all, the delay between a participant putting the armband on their forearm and the start of the recording was approximately five minutes.

The VR environment showed the participant the gesture to perform using an animation of a 3D arm. In place of their arm, the participants see a virtual prosthetic driven by the Leap Motion camera (as depicted by the yellow arm in Fig.~\ref{vr_interface}). Therefore, the participants received continuous feedback on the Leap Motion's perception of their arm. All gesture recordings were made with the participants standing up, with their forearm parallel to the floor, unsupported. Starting from the neutral gesture, they were instructed, with an auditory cue, to hold the depicted gesture for five seconds. The cue given to the participants was in the following form: ``Gesture X, 3, 2, 1, Go''. The data recording began just before the movement was started by the participant to capture the ramp-up segment of the muscle activity and always started with the neutral gesture. The recording of the eleven gestures for five seconds each was referred to as a \textit{cycle}. A total of four cycles (220s of data) were recorded with no interruption between cycles (unless requested by the participant). When recording the second cycle, the participants were asked to perform each gesture (except the neutral gesture) with maximum intensity. This second cycle serves as a baseline for the maximum intensity of each gesture (computed using the mean absolute value (MAV) of the EMG signal) on a given day, on top of providing labeled data. For the other three cycles, a natural level of intensity was requested from the participants. Note that while the participants interpreted the ``natural intensity command'' themselves, on average, the MAV of the signal recorded by each participant for each gesture during cycle 1, 3, and 4 was 43.43\%$\pm23.02\%$ of the MAV of the signal recorded during cycle 2 of the corresponding participant and gesture.

\subsubsection{Evaluation Session}

The evaluation session's main purpose was to generate data containing the four main dynamic factors within an online setting. The sessions took the form of a ``game'', where the participants were randomly requested to hold a gesture at a given intensity and position in 3D. Figure~\ref{vr_interface} provides an overview of the evaluation session.

The evaluation session always took place after a training session within the VR environment, without removing the armband between the two sessions. The participants were first asked to stand with their arm stretched forward to calibrate the user's maximum reach. Then, the participant was requested to bend their elbow 90 degrees, with their forearm parallel to the floor (this was the starting position). Once the participant is ready, the researcher starts the experiment which displays a countdown to the participant in the game. When the game starts, a random gesture is requested through text on the participant's head-up display. Additionally, a floating ring appears at a random position within reach of the participant, with a maximum angle of $\pm$45 and $\pm$70 degrees in pitch and yaw respectively. The floating ring's color (blue, yellow, or red) tells the participant at what level of intensity to perform the requested gesture. Three levels of intensity were used: (1) less than 25\%, (2) between 25 to 45\% and (3) above 45\% of the participant's maximal intensity as determined using cycle 2 of the requested gesture from the participant's first training session. The current gesture intensity of the participant was computed using the mean absolute value of the signal and a sliding window of 200 ms and updated every frame ($\sim$80 frames per second). A new gesture, position, and intensity are randomly requested every five seconds with a total of 42 gestures asked during an evaluation session (210 seconds).

During the experiment and using the Leap Motion, a \textit{virtual prosthetic arm} is mapped to the participant's arm, which matches its position and pitch/yaw angles. The participant thus intuitively knows where their arm is in the VR environment and how to best reach the floating ring. However, the virtual prosthetic does not match the participant's hand/wrist movements nor the forearm's roll angle. Instead, the Leap Motion's data is leveraged to predict the subject's current gesture using a convNet (see Section~\ref{leap_motion_convNet_section} for details). The hand of the virtual prosthetic then moves to perform the predicted gesture (including supination/pronation with the roll angle) based on the data recorded during the training session, providing direct feedback to the participant. Note that the sEMG data does not influence the gesture's prediction as to not bias the dataset toward a particular EMG classification algorithm. The virtual prosthetic also change color (blue, yellow, or red) based on the currently detected gesture intensity from the armband. Finally, a score is shown to the participant in real-time during the experiment. The score increases when the detected gesture matches the requested gesture. Bonus points are given when the participant correctly matches the requested gesture's position and intensity.

\subsection{Data Pre-processing}
\label{preprocessing_section}
This work aims at studying the effect of the four main dynamic factors in myoelectric control systems. Consequently, input latency is a critical factor to consider. As the optimal guidance latency was found to be between 150 and 250 ms~\cite{optimal_latency_real_time_EMG}, within this work, the EMG data from each participant is segmented into 150 ms frames with an overlap of 100 ms. The raw data is then band-pass filtered between 20-495 Hz using a fourth-order Butterworth filter. 

\subsection{Experiments with the Dynamic Dataset}

The training sessions will be used to compare the algorithms described in this work, in an offline setting. When using the training sessions for comparison, the classifiers will be trained on the first and third cycle and tested on the fourth cycle. The second cycle, comprised of the maximal intensity gestures recording, is omitted as to only take into account electrode shift/non-stationarity of the signal and to provide an easier comparison with the literature.

The evaluation session is employed to study the impact of the four main dynamic factors on EMG-based gesture recognition. Classifiers will again be trained on cycle 1 and 3 of the training sessions and tested on the two evaluation sessions.

\subsection{3DC Dataset}
A second dataset, referred to as the \textit{3DC Dataset} featuring 22 able-bodied participants, is used for architecture building, hyperparameter selection, and pre-training. This dataset presented in~\cite{3dc_armband}, features the same eleven gestures and is also recorded with the 3DC Armband. Its recording protocol closely matches the training session description (Section~\ref{training_session_section}), with the difference being that two such sessions were recorded for each participant (same day recording). This dataset was preprocessed as described in Section~\ref{preprocessing_section}. Note that when recording the 3DC Dataset, participants were wearing both the Myo and 3DC Armband. In this work, only the data from the 3DC Armband is considered.  


\section{Deep Learning Classifiers}
\label{deep_learning_classifiers}
The following section presents the deep learning architectures employed for the classification of both EMG data and images from the Leap Motion camera. The PyTorch~\cite{pytorch} implementation of the networks are \href{https://github.com/UlysseCoteAllard/LongTermEMG}{readily available here (https://github.com/UlysseCoteAllard/LongTermEMG)}.

\subsection{Leap Motion Convolutional Network}
\label{leap_motion_convNet_section}

For real-time myoelectric control, external feedback helps the participant adapt to improve the system's performance~\cite{visual_feedback_good_emg, journal_paper_TL_ulysse}. Such feedback is also natural to have as the participant should, in most cases, be able to see the effect of their actions. Consequently, to avoid biasing the proposed dataset toward a particular EMG-based classification algorithm, the gesture-feedback was provided using solely the Leap Motion. 

\begin{figure}[!htbp]
\centering
\includegraphics[width=.7\linewidth]{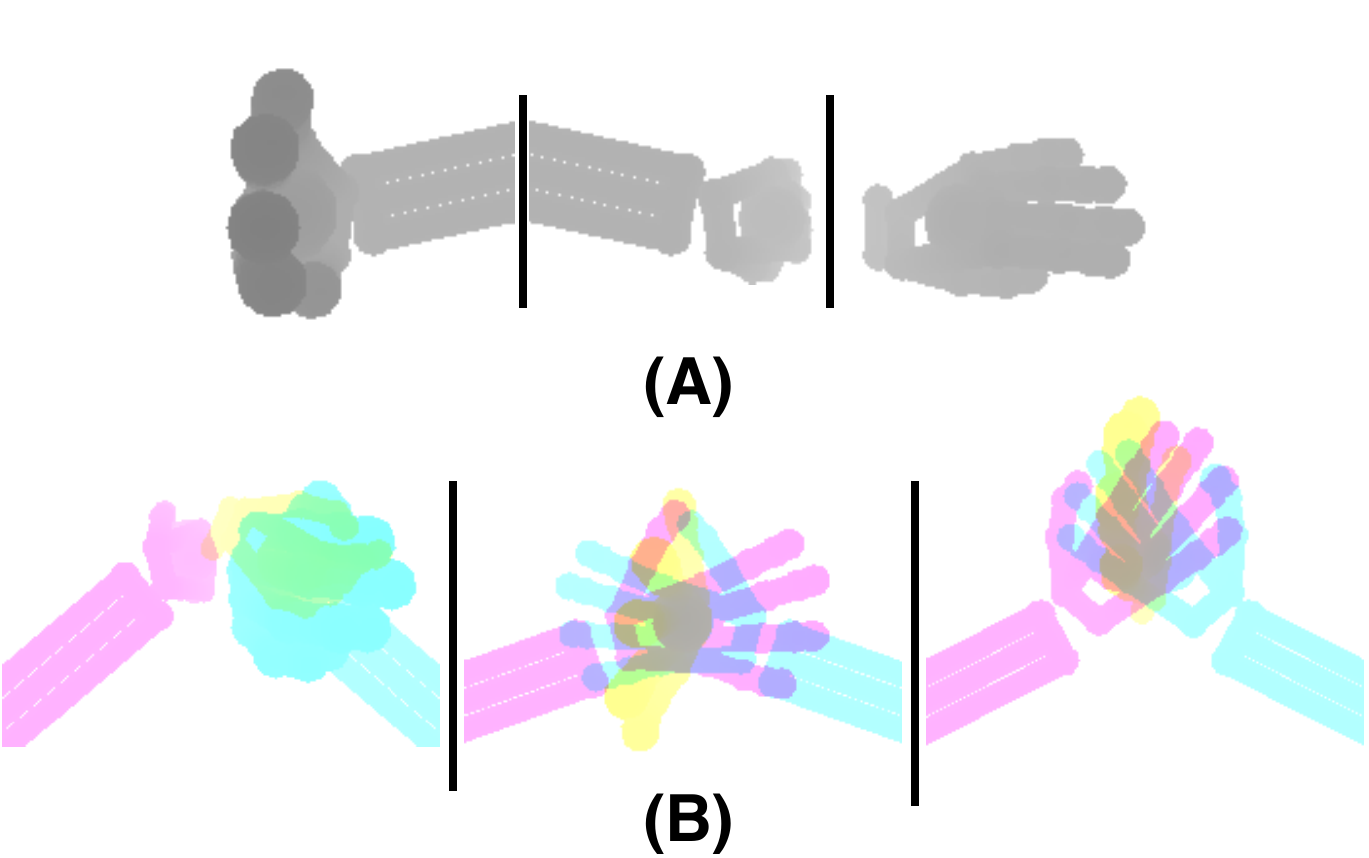}
\caption{(A) The depth images (darker pixels are closer) of the three virtual cameras taken at the same moment. The gesture captured is Wrist Flexion. Note that, regardless of the participant's movement, the three cameras are placed to have the same point-of-view relative to the forearm.
(B) Examples of images fed to the ConvNet. The represented gestures from left to right: Wrist Flexion, Open Hand, and Radial Deviation.}
\label{virtual_depth_camera}
\end{figure}

Image classification is arguably the domain in which ConvNet-based architecture had the greatest impact due, in part, to the vast amount of labeled data available~\cite{deep_learning}. However, within this work, and to provide consistent feedback, training data was limited to the first Training Session of each participant. Consequently, the network had to be trained with a low amount of data (around 200 examples per gesture). Additionally, while the training session was recorded with a fixed point of view of the participant's arm, the evaluation session, by design, introduced great variability in the limb's position, which the network had to contend with during inference. The variable point-of-view problem was addressed using the capability of the Leap Motion camera to generate a 3D model of the participant's hand in the virtual environment. Three virtual depth-cameras were then placed around the arm's 3D representation from three different and fixed points-of-view (relative to the participant's forearm) to capture images of the 3D model (see Figure~\ref{virtual_depth_camera}~(A) for an example). The three images were then merged by having each image encoded within one channel of a three-channel color image (see Figure~\ref{virtual_depth_camera}~(B) for examples). Finally, pixel intensity was inverted (so that a high value corresponds to a part of the hand being close to the camera) before being fed to the ConvNet. Note that one of the main reasons to utilize images as the input instead of a 3D point cloud is to reduce the computational requirements during both training and inference. 

The Leap Motion ConvNet's architecture is based on EfficientNet-B0~\cite{efficientNet} and presented in Table~\ref{LeapCameraConvNetArchitecture}.

\begin{table}[!ht]
\caption{Leap motion ConvNet's architecture}
\label{LeapCameraConvNetArchitecture}
\centering
\resizebox{.8\linewidth}{!}{
\begin{tabular}{cccccc}
\hline
\begin{tabular}[c]{@{}c@{}}Level\\ n\end{tabular} & Layer Type & \begin{tabular}[c]{@{}c@{}}Input Dimension\\ Height x Width\end{tabular} & \#Channels & \begin{tabular}[c]{@{}c@{}}\# Layers\\ Source Network\end{tabular} & \begin{tabular}[c]{@{}c@{}}\# Layers\\ Target Network\end{tabular} \\ \hline
1 & Conv3x3 & $225 \times 225$ & 33 & 1 & 1 \\
2 & ConvBlock3x3 & $113\times 113$ & 16 & 2 & 1 \\
3 & ConvBlock5x5 & $57\times 57$ & 24 & 2 & 1 \\
4 & ConvBlock3x3 & $29\times 29$ & 32 & 2 & 1 \\
5 & ConvBlock5x5 & $15\times 15$ & 48 & 2 & 1 \\
6 & ConvBlock5x5 & $8\times 8$ & 64 & 2 & 1 \\
7 & \begin{tabular}[c]{@{}c@{}}Conv1x1 \& \\ Pooling \&\\  FC\end{tabular} & $4\times 4$ & 64 & 1 & 1 \\ \hline
\end{tabular}}
\\Each row describes the level $n$ of the ConvNet. The pooling layer is a global average pooling layer (giving one value per channel), while ``FC'' refers to a fully connected layer.
\end{table}

\subsection{EMG-based Temporal Convolutional Network}

Temporal Convolutional Networks (TCN) generally differ in two aspects from standard ConvNets. First, TCNs leverage stacked layers of \textit{dilated convolutions} to achieve large receptive fields with few layers. Dilated convolutions (also known as convolution à trous or convolution with holes) is a convolutional layer where the kernel is applied over a longer range by skipping input values by a constant amount~\cite{wavenet}. Typically, the dilatation coefficient ($d$) is defined as $d=2^i$ where $i$ is the $i$th layer from the input (starting with i=0). The second difference is that TCNs are built with dilated \textit{causal} convolutions where the \textit{causal} part means that the output at time $t$ is convolved only with elements from outputs from time $t$ or earlier. In practice, such behavior is achieved (in the 1D case with PyTorch) by padding the left side (assuming time flows from left to right) of the vector to be convolved by $(k-1)*d$, where k is the kernel's size. This also ensures a constant output size throughout the layers. 

\begin{figure}[!htbp]
\centering
\includegraphics[width=\linewidth]{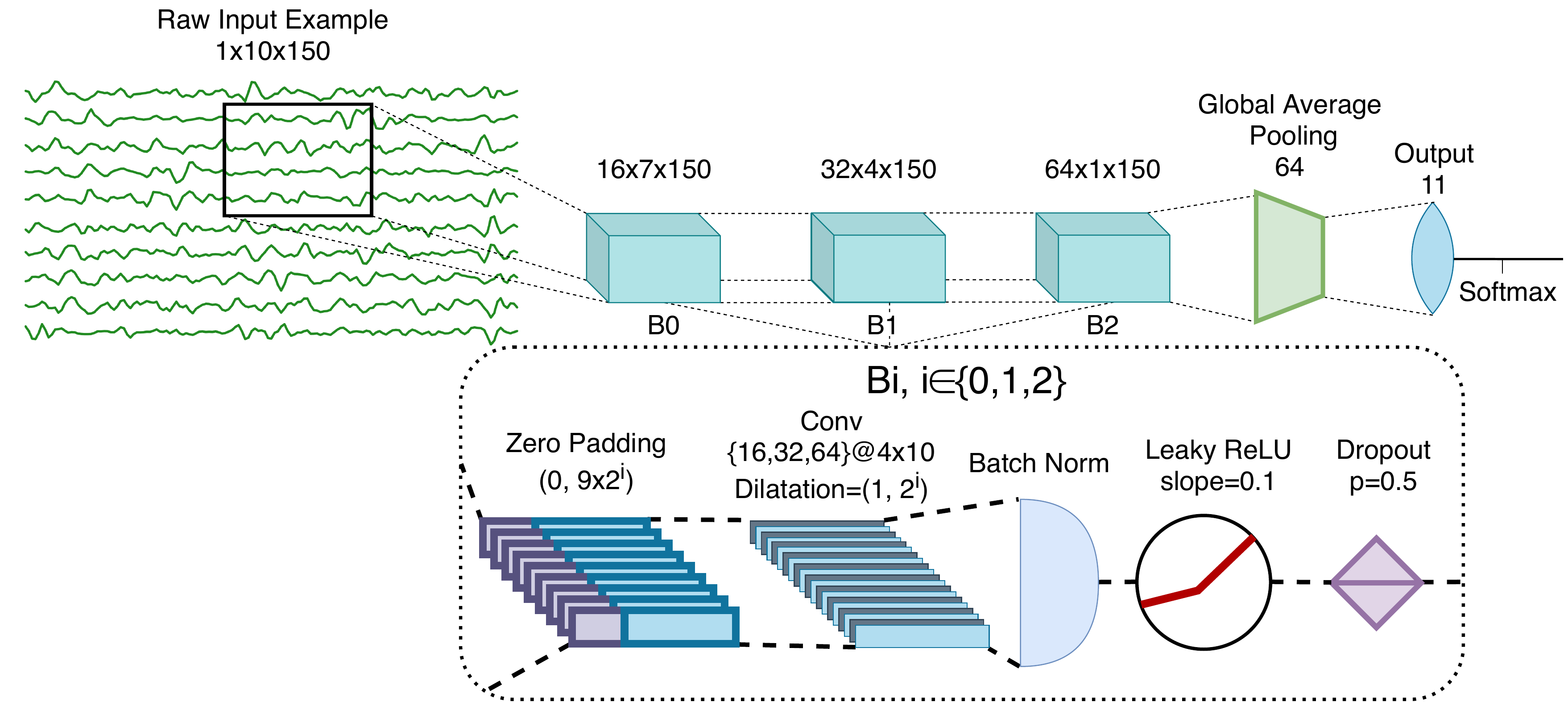}
\caption{The ConvNet's architecture employing 104 788 learnable parameters. In this figure, $B_i$ refers to the ith block ($i\in\{0,1,2\}$). Conv refers to a convolutional layer. The padding is removed after the convolution.}
\label{emgConvNetArchitecture}
\end{figure}

The proposed TCN, receives the sEMG data with shape Channel~$\times$~Time (10 $\times$ 150). The TCN's architecture (see Figure~\ref{emgConvNetArchitecture}) is based on~\cite{HandCraftVsLearnedFeaturesEMG, PyTorchTCN_implementation} and the PyTorch implementation is derived from~\cite{PyTorchTCN_implementation}. The network is made of three \textit{blocks} followed by a global average pooling layer before the output layer. Each block encapsulates a dilated causal convolutional layer~\cite{wavenet} followed by batch normalization~\cite{batch_normalization}, leaky ReLU~\cite{leaky_relu} and dropout~\cite{dropout}.

Adam~\cite{adam} is employed for the TCN's optimization with a batch size of 512. 10\% of the training data is held out as a validation set which is used for early stopping (with a ten epochs threshold) and learning rate annealing (factor of five and a patience of five). The learning rate (lr=0.002233) was selected by random search~\cite{random_search_bengio} using a uniform random distribution on a logarithmic scale between $10^{-5}$ and $10^1$ and 50 candidates (each candidate was evaluated 5 times). All architecture choices and hyperparameter selection were performed using the \textit{3DC Dataset}. 

\subsection{Calibration Training Methods}
This work considers three calibration methods for long-term classification of sEMG signals: No Calibration, Recalibration, and Delayed Calibration. In the first case, the network is trained solely from the data recorded during the first training session for each participant. In the Recalibration case, the model is re-trained for each participant with the data recorded from each new training session. To leverage previous data, fine-tuning~\cite{finetuning_networks} is applied. That is, during recalibration, the weights of the network are first initialized with the weights found from the previous training. Note that TADANN (see Section~\ref{transfer_learning}) will also use the Recalibration setting. Delayed Calibration is similar to Recalibration, but the network is recalibrated on the data recorded from the previous training session instead of the newest one. The purpose of Delayed Calibration is to evaluate the degradation in performance over the full experiment when using a classifier that is outdated by a constant amount of time at each session.




\section{Transfer Learning}
\label{transfer_learning}

Over multiple training sessions, a large amount of labeled data is recorded. However, standard training methods are limited to the data from the most recent session, as they cannot take into account the signal drift between each recording. Transfer learning algorithms on the other hand can be developed to account for such signal disparity. Consequently, this work proposes to combine the Adaptive Domain Adversarial Neural Network (ADANN) training presented in~\cite{HandCraftVsLearnedFeaturesEMG}, which tries to learn a domain-independent (in this case session-independent) feature space representation, with the transfer learning algorithm presented in~\cite{journal_paper_TL_ulysse} for inter-session gesture recognition. This new algorithm is referred to as Transferable Adaptive Domain Adversarial Neural Network (TADANN). For simplicity's sake, the ensemble of calibration sessions before the most recent one is referred to as the \textit{pre-calibration sessions}, whereas the most recent one is referred to as the \textit{calibration session}.

The proposed algorithm contains a pre-training (ADANN) and a training step. During pre-training, each session within the pre-calibration sessions is considered as a separate labeled domain dataset. At each epoch, pre-training is performed by sharing the weights of a network across all the domains (i.e. pre-calibration sessions), while the Batch-Normalization (BN) statistics are learned independently for each session~\cite{HandCraftVsLearnedFeaturesEMG}.
The idea behind ADANN is then to extract a general feature representation from this multi-domain setting. To do so, a domain classification head (with two neurons) is added to the network. At each epoch, for each step, a mini-batch is created containing examples from a single, randomly selected session at a time. A second mini-batch is then created from an also randomly selected session (different than the one used to create the source batch). The examples from the first mini-batch are assigned the domain-label 0, while the domain-label 1 is assigned to the examples from the second mini-batch. Then, a gradient reversal layer~\cite{DANN} is used right after the domain-head during backpropagation, to force the network to learn a session-independent feature representation. Note that the BN statistics used by the network correspond to the session from which the source or target batch originate, but that they are updated only with the source batch. Similarly, the classification head is used to backpropagate the loss only with the source batch. 

After pre-training, the learned weights are frozen, except for the BN parameters, which allows the network to adapt to a new session. Then, a second network is initialized (in this work, the second network is identical to the pre-trained network) and connected with an element-wise summation operation in a layer-by-layer fashion to the pre-trained network (see~\cite{journal_paper_TL_ulysse} for details). Additionally, all outputs from the pre-trained network are multiplied by a learnable coefficient (clamped between 0 and 2) before the summation, as to provide an easy mechanism to neuter or increase the influence of the pre-trained network at a layer-wise level.

\section{Results}
\label{results}

\subsection{Training Sessions: Over time classification accuracy}
\begin{figure}[!htbp]
\centering
\includegraphics[width=\linewidth]{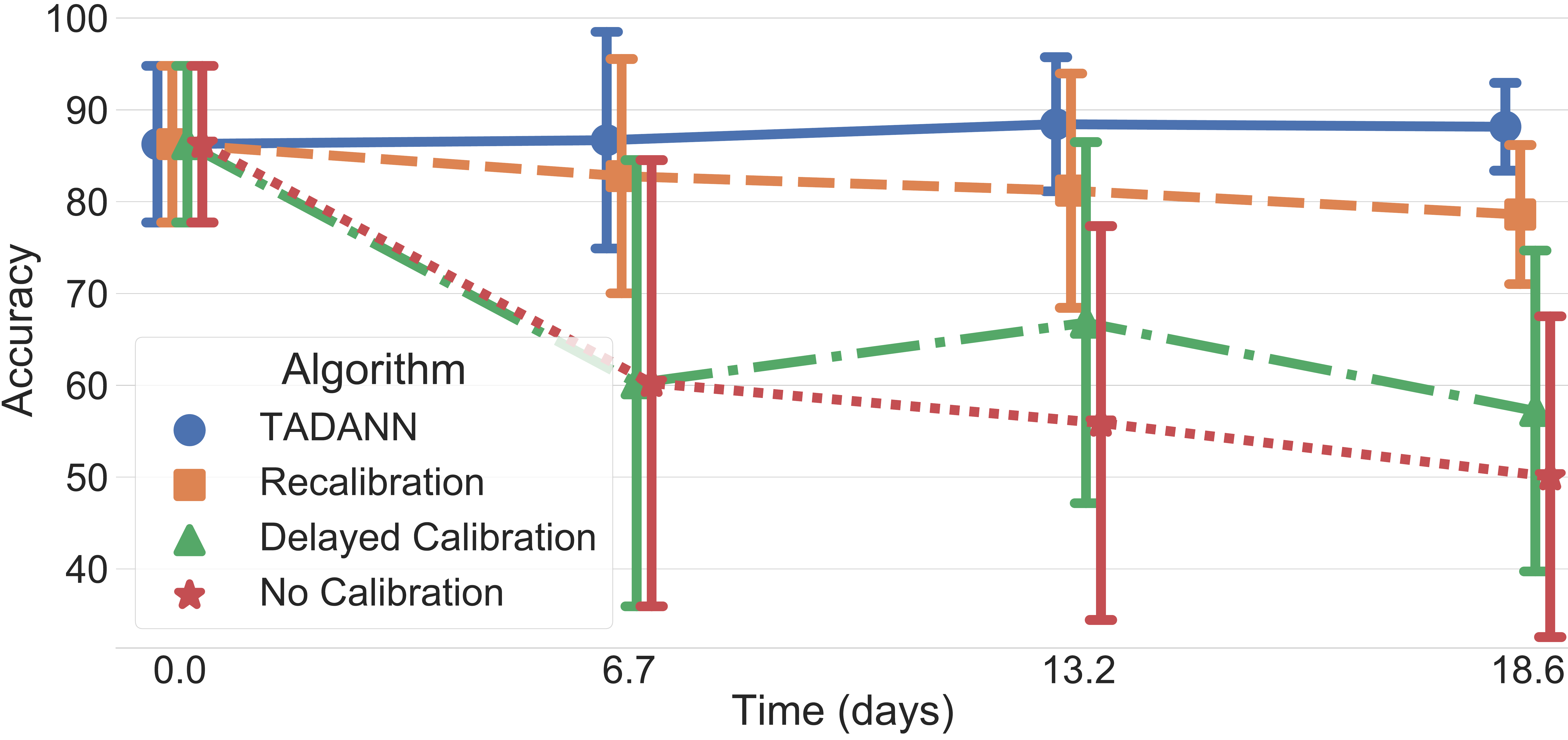}
\caption{Long-term EMG Classification on Training Session Experiment: Average accuracy over time calculated on the last cycle of the training sessions (classifiers trained on cycle one and three). The values given on the x-axis represent the average time (in days) elapsed between the current session and the first session across all participants.}
\label{accuracy_over_time_training_session}
\end{figure}

Figure~\ref{accuracy_over_time_training_session} shows the average accuracy over time across all participants for TADANN, Recalibration, Delayed Calibration, and No Calibration.

Table~\ref{table:offline_accuracy} presents the average accuracy across all participants (classifier trained on cycle one and three) for each session for the Recalibration, TADANN, No Calibration and Delayed Calibration training scheme. Additionally, Cohen's $D_z$~\cite{cohenD, cohens_dz_explanation_and_relevance} is given to help determine the effect size by providing a measure of standardized mean difference between using the fine-tuning recalibration method or the other recalibration schemes. Conventionally, the effect size is referred to as small (=0.2), medium (=0.5), or large (=0.8). It is important to note however that these values should not be interpreted rigidly~\cite{cohens_dz_explanation_and_relevance}. Using the Wilcoxon signed-rank test~\cite{use_friedman_plus_holm} shows that TADANN significantly outperforms Recalibration (p-value $=0.00194$, $0.00025$ and $0.02771$ for session two (n=20), three (n=20) and four (n=6) respectively).

\begin{table}[!ht]
\caption{Offline accuracy for eleven gestures over multiple days}
\centering
\label{table:offline_accuracy}
\resizebox{.85\linewidth}{!}{
\begin{tabular}{@{}ccccc@{}}
\toprule
 &
  Recalibration &
  TADANN &
  No Calibration &
  Delayed Calibration \\ \midrule
\begin{tabular}[c]{@{}c@{}}Session 1\\ STD\end{tabular} &
  \begin{tabular}[c]{@{}c@{}}N\textbackslash{}A\\ N\textbackslash{}A\end{tabular} &
  \begin{tabular}[c]{@{}c@{}}N\textbackslash{}A\\ N\textbackslash{}A\end{tabular} &
  \begin{tabular}[c]{@{}c@{}}86.26\%\\ 8.75\%\end{tabular} &
  \begin{tabular}[c]{@{}c@{}}N\textbackslash{}A\\ N\textbackslash{}A\end{tabular} \\ \midrule
\begin{tabular}[c]{@{}c@{}}Session 2\\ STD\\ Cohen's Dz\end{tabular} &
  \begin{tabular}[c]{@{}c@{}}82.78\%\\ 13.10\%\\ N\textbackslash{}A\end{tabular} &
  \textbf{\begin{tabular}[c]{@{}c@{}}86.70\%\\ 12.11\%\\ 0.58\end{tabular}} &
  \begin{tabular}[c]{@{}c@{}}60.22\%\\ 24.94\%\\ -1.23\end{tabular} &
  \begin{tabular}[c]{@{}c@{}}N\textbackslash{}A\\ N\textbackslash{}A\\ N\textbackslash{}A\end{tabular} \\ \midrule
\begin{tabular}[c]{@{}c@{}}Session 3\\ STD\\ Cohen's Dz\end{tabular} &
  \begin{tabular}[c]{@{}c@{}}81.20\%\\ 13.09\%\\ N\textbackslash{}A\end{tabular} &
  \textbf{\begin{tabular}[c]{@{}c@{}}88.44\%\\ 7.50\%\\ 0.83\end{tabular}} &
  \begin{tabular}[c]{@{}c@{}}55.89\%\\ 22.01\%\\ -1.51\end{tabular} &
  \begin{tabular}[c]{@{}c@{}}66.82\%\\ 20.18\%\\ -0.78\end{tabular} \\ \midrule
\begin{tabular}[c]{@{}c@{}}Session 4\\ STD\\ Cohen's Dz\end{tabular} &
  \begin{tabular}[c]{@{}c@{}}78.60\%\\ 8.30\%\\ N\textbackslash{}A\end{tabular} &
  \textbf{\begin{tabular}[c]{@{}c@{}}88.16\%\\ 5.24\%\\ 2.02\end{tabular}} &
  \begin{tabular}[c]{@{}c@{}}50.05\%\\ 19.14\%\\ -1.43\end{tabular} &
  \begin{tabular}[c]{@{}c@{}}57.20\%\\ 19.14\%\\ -0.93\end{tabular} \\ \bottomrule
\end{tabular}
}
\end{table}


\subsection{Evaluation Session}
\begin{figure}[!htbp]
\centering
\includegraphics[ width=.9\linewidth]{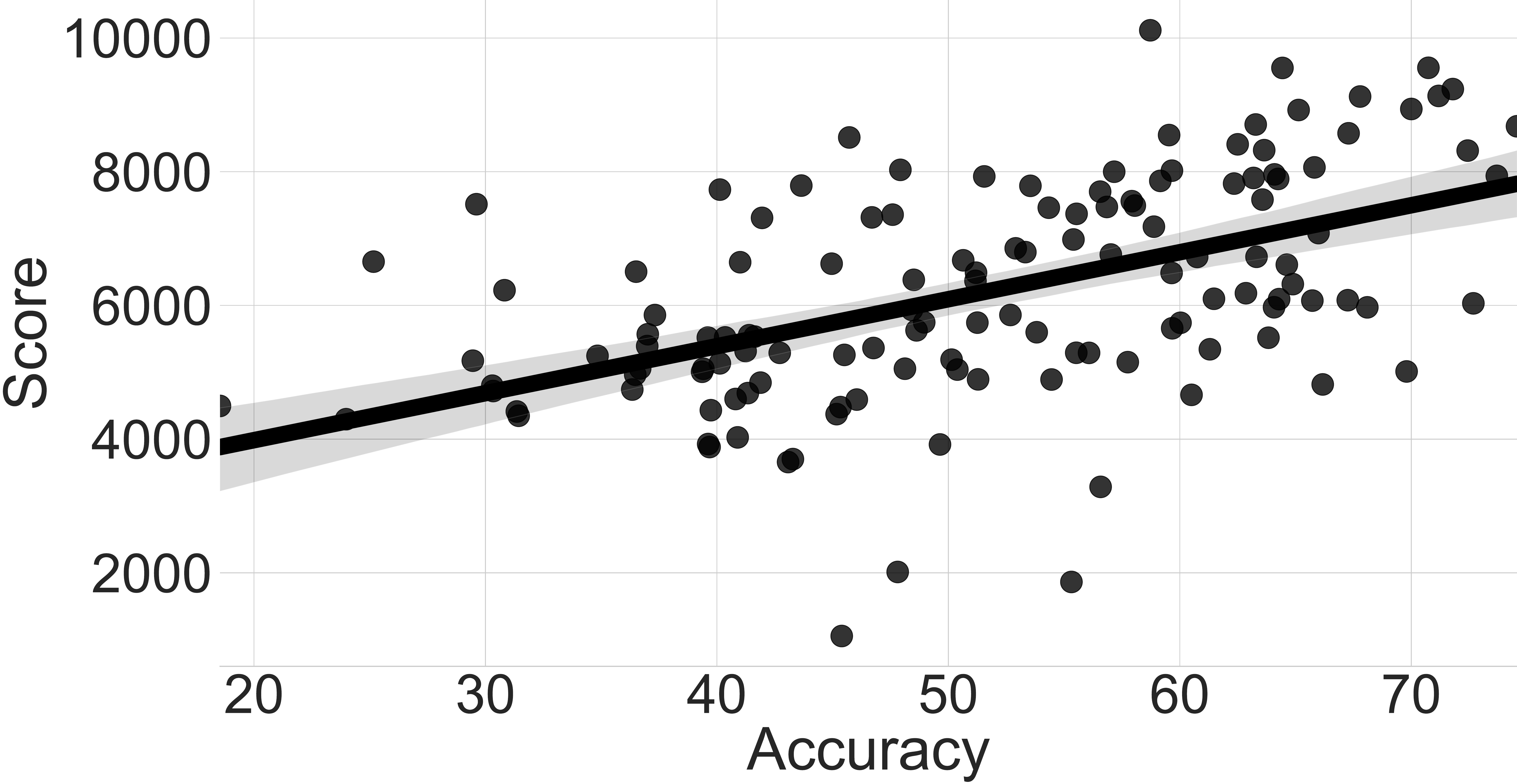}
\caption{The score obtained by each participant at each evaluation session with respect to TADANN's accuracy on the evaluation sessions. The translucent bar around the regression represents the standard deviation. The Pearson r correlation coefficient between the score and accuracy is 0.52}
\label{score_vs_accuracy_correlation_figure}
\end{figure}

Figure~\ref{score_vs_accuracy_correlation_figure} shows the scores obtained for all participants on the evaluation sessions with respect to TADANN's accuracy from the corresponding session. The Pearson r correlation coefficient between the score and accuracy is $0.52$. The average score obtained during the first recording session was 5634$\pm$1521, which increased to 6615$\pm$1661 on session three, showing that the participants improved over time.

\subsubsection{Over time classification accuracy}

\begin{figure}[!htbp]
\centering
\includegraphics[width=\linewidth]{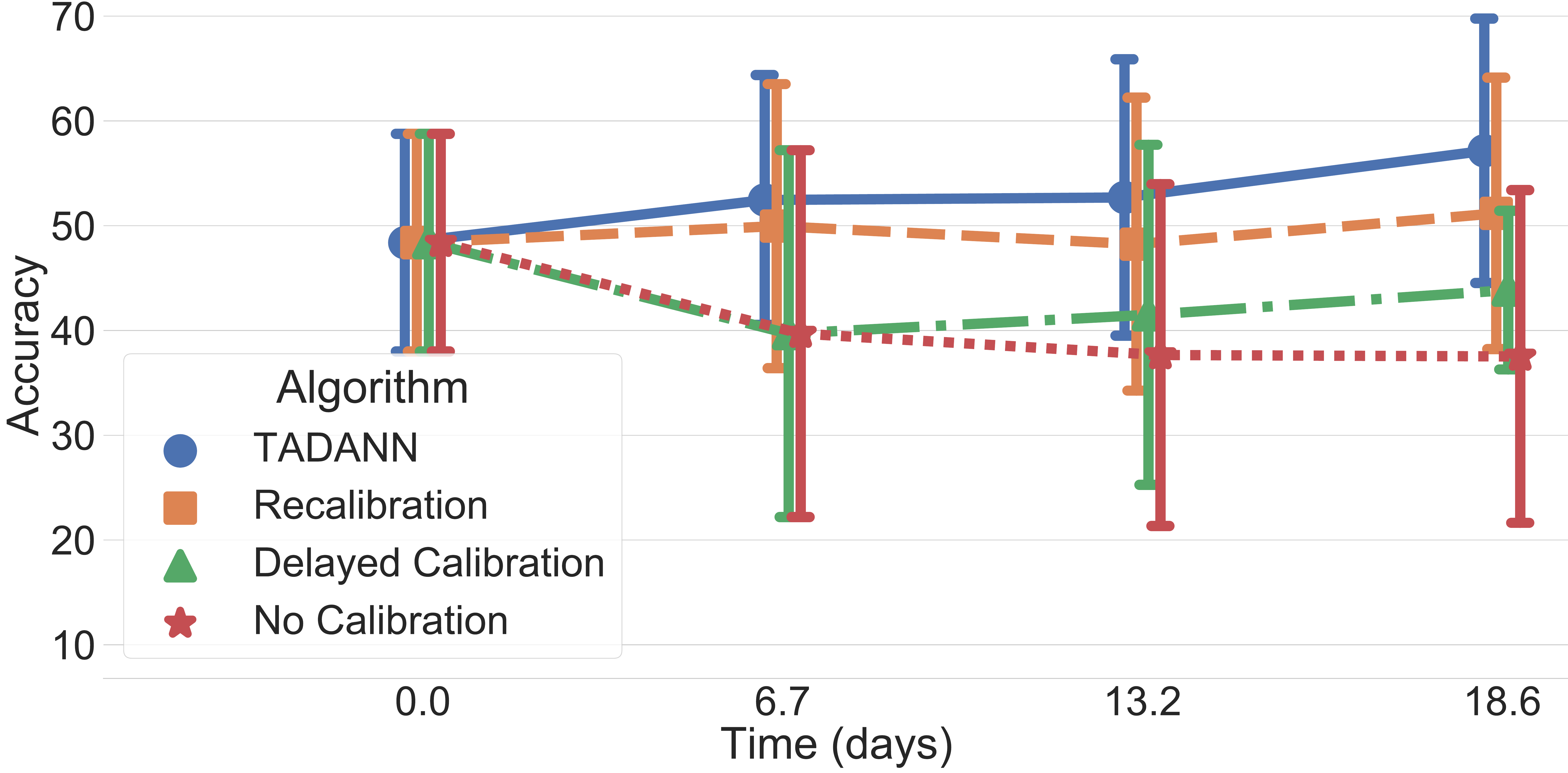}
\caption{Long-term EMG Classification on Evaluation Session Experiment: Average accuracy over time on the evaluation sessions (classifiers trained on cycle one and three). The values given on the x-axis represent the average time (in days) elapsed between the current session and the first session across all participants.}
\label{accuracy_overtime_online_with_transition}
\end{figure}

Figure~\ref{accuracy_overtime_online_with_transition} shows the average accuracy over time on the evaluation sessions, across all participants for TADANN, Recalibration, Delayed Calibration, and No Calibration.

Table~\ref{table:online_accuracy} presents the average accuracy for each session on the Evaluation Sessions across all participants (classifier trained on cycle one and three) for all calibration methods. TADANN again significantly outperformed the Recalibration scheme using the Wilcoxon signed-rank test (p-value $=0.00194$, $0.002204$ and $0.04640$ for session two (n=20), three (n=20) and four (n=6) respectively.

\begin{table}[!ht]
\caption{Online accuracy for eleven gestures over multiple days}
\centering
\label{table:online_accuracy}
\resizebox{.85\linewidth}{!}{
\begin{tabular}{ccccc}
\hline
 &
  Recalibration &
  TADANN &
  No Calibration &
  Delayed Calibration \\ \hline
\begin{tabular}[c]{@{}c@{}}Session 1\\ STD\end{tabular} &
  \begin{tabular}[c]{@{}c@{}}N\textbackslash{}A\\ N\textbackslash{}A\end{tabular} &
  \begin{tabular}[c]{@{}c@{}}N\textbackslash{}A\\ N\textbackslash{}A\end{tabular} &
  \begin{tabular}[c]{@{}c@{}}48.38\%\\ 10.04\%\end{tabular} &
  \begin{tabular}[c]{@{}c@{}}N\textbackslash{}A\\ N\textbackslash{}A\end{tabular} \\ \hline
\begin{tabular}[c]{@{}c@{}}Session 2\\ STD\\ Cohen's Dz\end{tabular} &
  \begin{tabular}[c]{@{}c@{}}49.96\%\\ 13.65\%\\ N\textbackslash{}A\end{tabular} &
  \textbf{\begin{tabular}[c]{@{}c@{}}52.45\%\\ 11.96\%\\ 0.66\end{tabular}} &
  \begin{tabular}[c]{@{}c@{}}39.70\%\\ 17.70\%\\ -1.22\end{tabular} &
  \begin{tabular}[c]{@{}c@{}}N\textbackslash{}A\\ N\textbackslash{}A\\ N\textbackslash{}A\end{tabular} \\ \hline
\begin{tabular}[c]{@{}c@{}}Session 3\\ STD\\ Cohen's Dz\end{tabular} &
  \begin{tabular}[c]{@{}c@{}}48.24\%\\ 13.19\%\\ N\textbackslash{}A\end{tabular} &
  \textbf{\begin{tabular}[c]{@{}c@{}}52.69\%\\ 12.45\%\\ 0.79\end{tabular}} &
  \begin{tabular}[c]{@{}c@{}}37.66\%\\ 16.37\%\\ -1.27\end{tabular} &
  \begin{tabular}[c]{@{}c@{}}41.49\%\\ 16.02\%\\ -0.75\end{tabular} \\ \hline
\begin{tabular}[c]{@{}c@{}}Session 4\\ STD\\ Cohen's Dz\end{tabular} &
  \begin{tabular}[c]{@{}c@{}}51.18\%\\ 13.96\%\\ N\textbackslash{}A\end{tabular} &
  \textbf{\begin{tabular}[c]{@{}c@{}}57.14\%\\ 13.59\%\\ 1.19\end{tabular}} &
  \begin{tabular}[c]{@{}c@{}}37.52\%\\ 17.16\%\\ -0.95\end{tabular} &
  \begin{tabular}[c]{@{}c@{}}43.85\%\\ 7.82\%\\ -0.63\end{tabular} \\ \hline
\end{tabular}
}
\end{table}

\subsubsection{Limb orientation}

\begin{figure}[!htbp]
\centering
\includegraphics[trim={10cm 12cm 10cm 12cm}, clip, width=.8\linewidth]{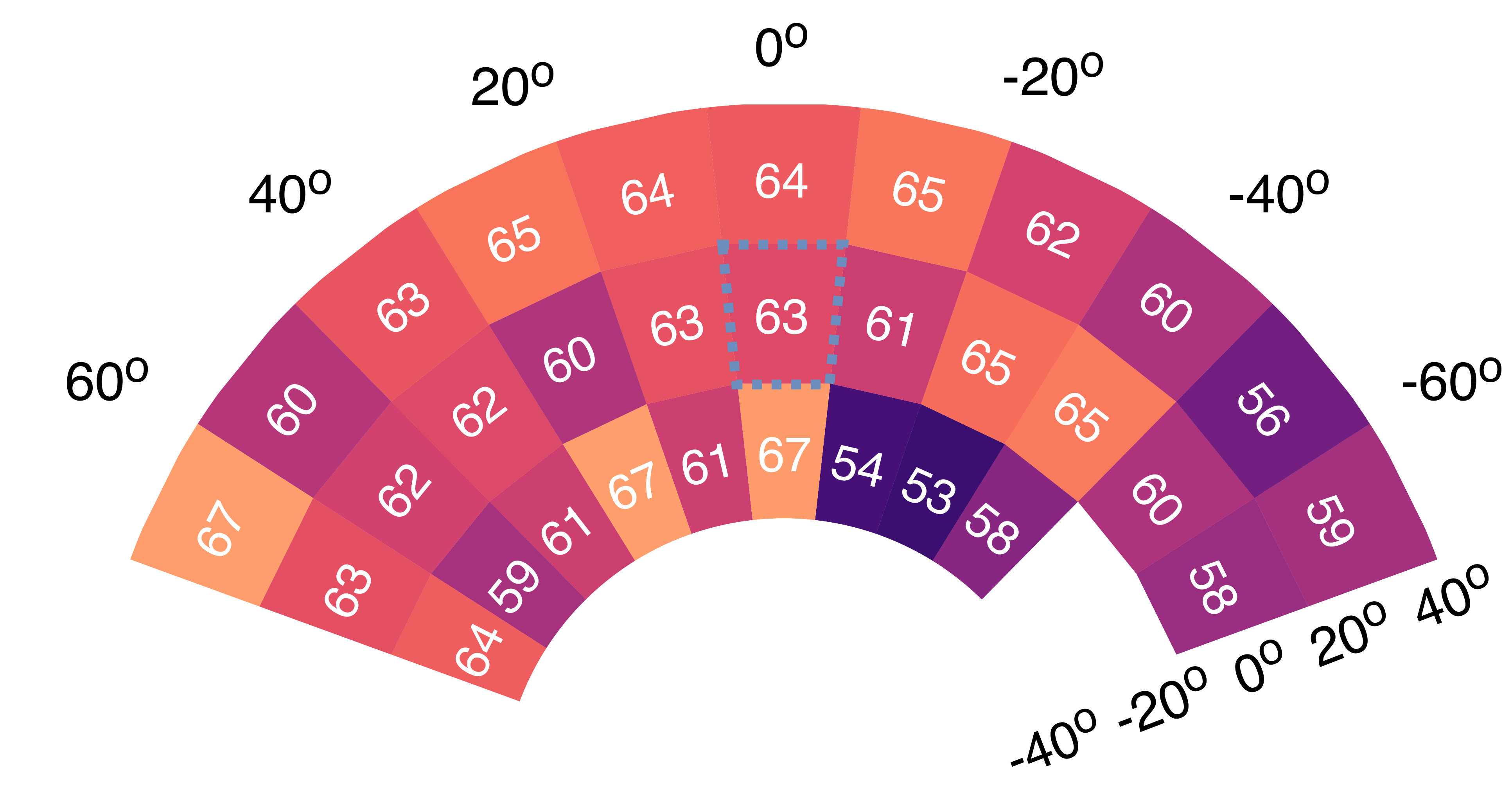}
\caption{Limb Position Experiment: Accuracy with respect to the pitch and yaw angles. The dotted line indicates the neutral orientation. Note that, a minimum threshold of 500 examples per pitch/yaw combination was set to show the accuracy.}
\label{no_transition_angles_figure}
\end{figure}

The impact of the limb's position on the Recalibrated ConvNet's accuracy is shown in Figure~\ref{no_transition_angles_figure}. Accuracies were computed on the online dataset across all sessions and all participants. The first 1.5s after a new gesture was requested were removed from the data used to generate Figure~\ref{no_transition_angles_figure}, to reduce the impact of the gesture transition. 

\subsubsection{Gesture intensity}
\begin{figure}[!htbp]
\centering
\includegraphics[width=\linewidth]{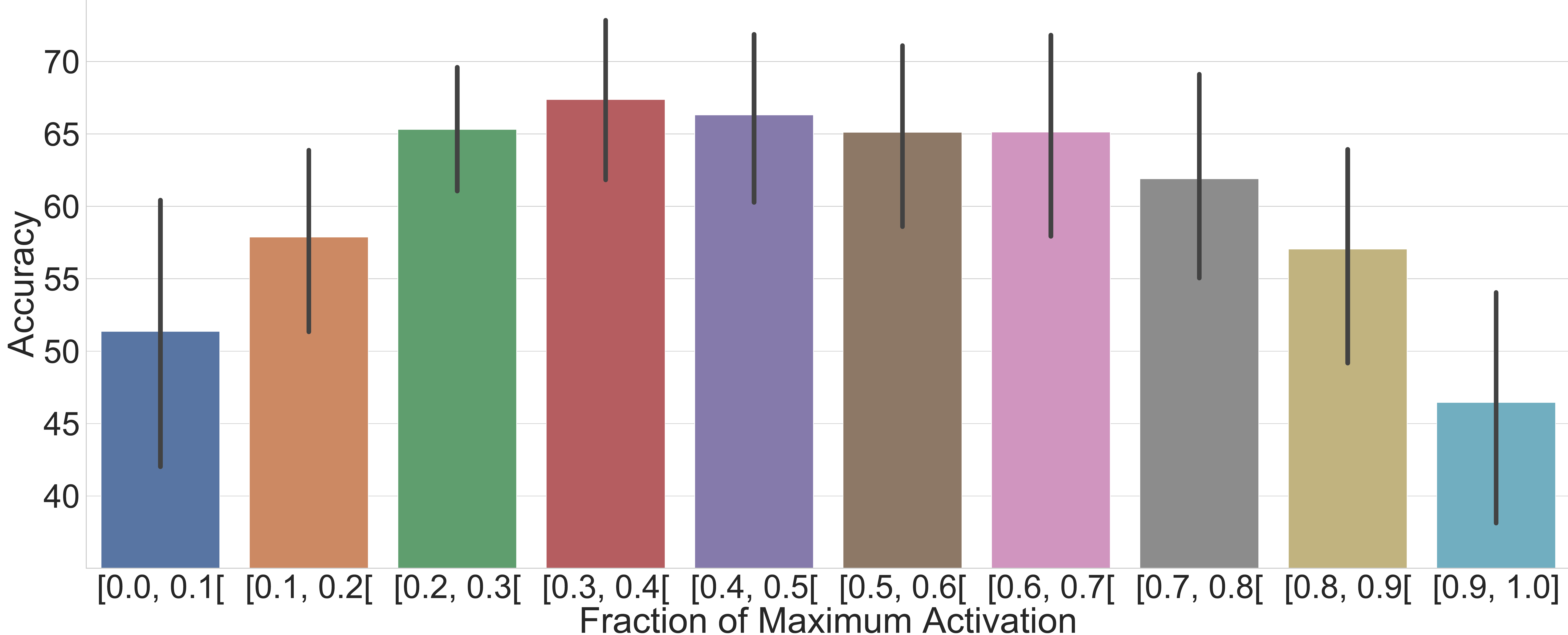}
\caption{Gesture Intensity Experiment: Average accuracy obtained from the recalibrated ConvNet with respect to the percentage of the maximum myoelectric activity when performing the gestures over all evaluation sessions across all participants.}
\label{muscle_intensity_graph}
\end{figure}

Figure~\ref{muscle_intensity_graph} shows the impact of the gesture's intensity on the Recalibration classifier's accuracy. Accuracies were computed on the online dataset across all sessions and all participants (excluding the neutral gesture). The first 1.5s after a new gesture was requested were again removed from the data to reduce the impact of gesture transition.

\section{Discussion}
\label{discussion}

This paper leverages the Leap Motion for gesture recognition, to avoid biasing the dynamic dataset toward a particular sEMG-based algorithm. Figure~\ref{score_vs_accuracy_correlation_figure} shows that the score obtained from a session correlates with the accuracy obtained from the same session. Note that the three lowest scores come from sessions where the Leap Motion lost tracking of the hand particularly often. Qualitatively, the participants enjoyed the experiment gamification as almost all of them were trying to beat their high-score and to climb to the top of the leaderboard. Additionally, several participants requested to do ``one more try'' to attempt to achieve a new high-score (only authorized after their last session). As such, virtual reality can provide an entertaining environment, from which to perform complex 3D tasks~\cite{emg_vr_2016} at an affordable cost when compared to the use of robotic arms or myoelectric prosthesis. 

The inter-day classification was shown to have a substantial impact on both offline and online. With standard classification algorithms, the need for periodic recalibration is thus apparent. The proposed TADANN algorithm was shown to consistently achieve higher accuracy than using fine-tuning to recalibrate the network. In this particular dataset, on a per-subject basis, TADANN routinely outperformed fine-tuning by more than 5\%, whereas for the opposite 1\% or less was the most common. The difference between the two also grew as TADANN could pre-train on more sessions. Thus future work will consider even more sessions per participant to evaluate TADANN.

\subsection{Controller feedback and user adaptation}
\begin{figure}[!htbp]
\centering
\adjincludegraphics[width=\linewidth, trim={0 0 0 {.45\height}}]{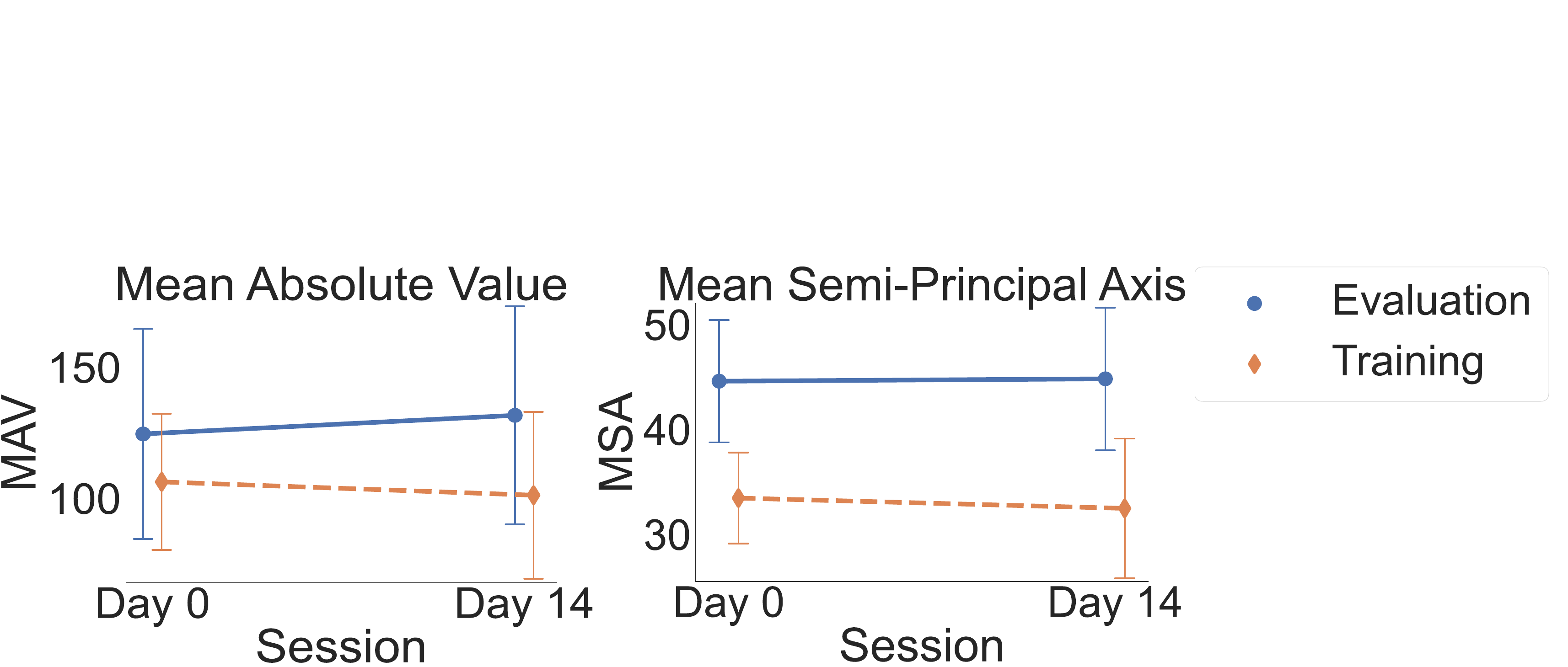}
\caption{Comparison between Day 0 and Day 14 of the MAV and MSA averaged across all participants. Error bars show the standard deviation. Cycle 2 (maximal intensity) of the training session is excluded from the computation. Note that both the MAV and MSA are higher during the Evaluation Sessions than the Training Sessions. This is expected as the Evaluation Sessions includes variation both in limb position and gesture intensity, which is not the case for the Training Sessions.}
\label{fig:mav_and_msa}
\end{figure}

Comparing the Delayed Calibration with the No Calibration from Figure~\ref{accuracy_over_time_training_session} and~\ref{accuracy_overtime_online_with_transition} show that participants were able to learn to produce more easily distinguishable gestures across sessions (from an EMG-based classifier perspective) without receiving feedback from an EMG-based controller. In a user-training study, Kristoffersen et al.~\cite{feedback_vs_no_feedback_affect_EMG} have shown that when compared with EMG-based continuous feedback, training with no feedback increases the amplitude (measured by the MAV) and pattern variability (measured by the mean semi-principal axis (MSA)~\cite{msa, msa_pca}) of the EMG signal over time (measured over several days). In this study, however, the gesture's feedback was based on the Leap Motion. Further, continuous feedback was only provided during the Training Sessions, as during the Evaluation Sessions, the feedback provided came from the Leap Motion-based classifier. Therefore, it is important to see if, within the dynamic dataset, the participants' behavior was more akin to the no feedback or the EMG-based continuous feedback. Fig.~\ref{fig:mav_and_msa} shows both the MAV and MSA computed on day 0 and day 14 for both the Training and Evaluation Sessions, averaged across all participants. For both metrics and session types, there is no significant difference between day 0 and day 14 using the Wilcoxon Signed-Rank test~\cite{use_friedman_plus_holm}. This would indicate that the user's adaptation within the proposed VR environment generates EMG signals whose properties over time match more closely to the type of EMG signal (for amplitude and pattern variability) generated when the user is receiving continuous feedback. Thus, the feedback provided by the Leap Motion seems to act as a good proxy, while also removing the bias normally present in online datasets.  

\subsection{Limb Orientation}
Figure~\ref{no_transition_angles_figure} shows that gestures that were performed while the participant's arm was externally rotated were the hardest in general for the classifier to correctly predict. This is likely due to the fact that the origin of the brachioradialis muscle (which is under the area of recording) is the lateral supracondylar ridge of the humerus. It is possible, therefore, that as the humerus becomes more externally rotated that it changes the geometry of the brachioradialis, affecting the observed signals. In addition, the arm may tend to supinate slightly for higher levels of external humeral rotation, which is known to create a worse limb position effect than the overall arm position. In contrast, when the participant's arm was internally rotated, no such drastic drop in performance was noted. As shown in~\cite{limb_position_erik}, training a classifier by including multiple limb-positions can improve inter-position performances. Consequently, it might be beneficial for future studies to focus on yaw-rotated forearm positions within the training dataset. Note, however, that while the participants were instructed to limit any torso rotation as much as possible, they were not restrained and consequently such rotation is likely present within the dataset. This might explain the decrease$\rightarrow$increase$\rightarrow$decrease in accuracy observed for the external rotation. Participants accepted an external rotation up to when they felt uncomfortable and then rotated their torso. This also explains the lower number of examples with an external yaw and a downward pitch, as such combinations tend to be uncomfortable (the software considered all angle combination with equal probability). 

\subsection{Gesture Intensity}
The impact of the gesture's intensity obtained within this study corroborates past findings in the literature~\cite{transcientChangeEMG_including_gesture_intensity}.
The classifier is relatively unaffected by different levels of gesture intensity between 20 and 70\%. Additionally, at lower intensity, the main error factor comes from classifying the neutral gesture. However, it has been shown that rejection-based classifiers can improve controller's usability~\cite{rejection_based_classification_good}. The problematic intensities are thus all above 70\% of maximal gesture intensity.

\subsection{Limitations}
The main limitation of this study is the relatively important gap between sessions. While such a scenario is realistic (e.g. for a consumer-grade armband used to play video games~\cite{serious_game_rehabilitation} or artistic performances~\cite{dance_journal_paper}) it is not possible to smoothly evaluate the change in signals within days. As such, future works will expand upon the current dataset to include more frequent evaluation sessions for each participant (and multiple within the same day). 

Another limitation of this work is that it only considers discrete gesture feedback during the Evaluation Sessions. Within the context of myoelectric control, it is easier for the user to positively adapt (i.e. improve system usability) when using a continuous (regressor) controller compared to a discrete (classifier) one~\cite{user_adaptation_discrete_vs_continuous}. What is unclear however, is how the user learning on a controller using a different modality (e.g. stereo-images) affects the performance of the EMG-based controller and if the type of controller (discrete or continuous) has an impact on this cross-modality adaptation. These questions, alongside the usability of the proposed dynamic dataset as a benchmark for EMG-based regressors, will be explored in future works.

\section{Conclusion}

This paper presents a new VR experimental protocol for sEMG-based gesture recognition, leveraging the Leap Motion camera so as to not bias the online dataset. Quantitatively and qualitatively, the participants were shown to improve over time and were motivated in taking part in the experiment. Overall, TADANN was shown to significantly outperform fine-tuning. The VR environment in conjunction with the Leap Motion allowed for the quantification of the impact of limb position with, to the best of the authors' knowledge, the highest resolution yet.

\section*{Acknowledgment}
The authors would like to thank Alexandre Campeau-Lecours for his support, without which this manuscript would not have been possible.

\section*{Funding}
This research was funded by the Natural Sciences and Engineering Research Council of Canada (NSERC)
[funding reference numbers 401220434, 376091307, 114090, 201404920], the Institut de recherche Robert-Sauvé en sante et en sécurité du travail (IRSST), and the Research Council of Norway under Grant 262762 and Grant 274996. Cette recherche a été financée par le Conseil de recherches en sciences naturelles et en génie du Canada (CRSNG) [numéros de références 401220434, 376091307, 114090, 201404920].

\ifCLASSOPTIONcaptionsoff
  \newpage
\fi



%
\bibliographystyle{IEEEtran}
\bibliography{bibliography}

\end{document}